\documentclass[letterpaper]{article} %
\usepackage{aaai25}  %
\usepackage{times}  %
\usepackage{helvet}  %
\usepackage{courier}  %
\usepackage[hyphens]{url}  %
\usepackage{graphicx} %
\usepackage{natbib}  %
\usepackage{caption} %
\usepackage{algorithm}
\usepackage{algorithmic}

\usepackage{newfloat}
\usepackage{listings}
\DeclareCaptionStyle{ruled}{labelfont=normalfont,labelsep=colon,strut=off} %
\floatstyle{ruled}
\newfloat{listing}{tb}{lst}{}
\floatname{listing}{Listing}
\title{MAGIC: Generating Self-Correction Guideline for In-Context Text-to-SQL}
\author{
    Arian Askari\textsuperscript{\rm 1}\thanks{\ \ Work done during an internship at Microsoft.}, Christian Poelitz\textsuperscript{\rm 2}, Xinye Tang\textsuperscript{\rm 3}
}
\affiliations{
    \textsuperscript{\rm 1}Leiden University 
    a.askari@liacs.leidenuniv.nl\\
    \textsuperscript{\rm 2}Microsoft Research Cambridge, UK, 
    \textsuperscript{\rm 3}Microsoft Redmond 
    \{cpoelitz, xinye.tang\}@microsoft.com
}

\usepackage{bibentry}

\usepackage{multirow}
\usepackage{booktabs} %
\usepackage{times}
\usepackage{mdframed}
\usepackage{latexsym}
\usepackage[inline]{enumitem}
\usepackage{tikz}
\usepackage{verbatimbox}
\usepackage{pdfpages}
\usepackage{amsmath}
\mdfdefinestyle{roundedstyle}{
    backgroundcolor=verylightgray,
    roundcorner=10pt,
    innertopmargin=10pt,
    innerbottommargin=10pt,
    innerleftmargin=10pt,
    innerrightmargin=10pt,
    linewidth=1pt
}
\newcommand{\header}[1]{\vspace*{1mm}\noindent\textbf{#1}.}
\usepackage{xcolor}
\definecolor{verylightgray}{gray}{0.95} %
\usepackage[T1]{fontenc}
\usepackage[utf8]{inputenc}
\usepackage{microtype}
\usepackage{inconsolata}
\usepackage{graphicx,verbatimbox}

\begin{document}

\maketitle

\begin{abstract}
Self-correction in text-to-SQL is the process of prompting large language model (LLM) to revise its previously incorrectly generated SQL, and commonly relies on manually crafted self-correction guidelines by human experts that are not only labor-intensive to produce but also limited by the human ability in identifying all potential error patterns in LLM responses. We introduce MAGIC, a novel \textbf{m}ulti-\textbf{ag}ent method that automates the creat\textbf{i}on of the self-\textbf{c}orrection guideline. MAGIC uses three specialized agents: a manager, a correction, and a feedback agent. These agents collaborate on the failures of an LLM-based method on the training set to iteratively generate and refine a self-correction guideline tailored to LLM mistakes, mirroring human processes but without human involvement. Our extensive experiments show that MAGIC's guideline outperforms expert human's created ones. We empirically find out that the guideline produced by MAGIC enhances the interpretability of the corrections made, providing insights in analyzing the reason behind the failures and successes of LLMs in self-correction. 
All agent interactions are publicly available at \url{https://huggingface.co/datasets/microsoft/MAGIC}.
\end{abstract}

\section{Introduction}
Converting natural language questions to SQL database queries, known as text-to-SQL, serves as a pivotal component for empowering non-expert data analysts in extracting desired information from relational databases using natural language \cite{qu2024before}.    
While large language models have shown a significant improvement in text-to-SQL and serve as state-of-the methods according to the leaderboards \cite{leaderboardBIRD}, they are prone to mistakes and even GPT4 has a notable accuracy gap of 30\% within human \cite{li2023can}.  
\par
A solution for resolving the mistakes of LLM is the emerging concept of `self-correction' \cite{madaan2023self} that defines as the ability of LLMs in revising their previous mistakes under the hypothesis that recognizing errors is easier than avoiding them \cite{gou2024critic,madaan2023self}. 
Self-correction in context of text-to-SQL is the process of prompting an LLM to revise its previously incorrectly generated SQL \cite{zhang2024benchmarking,chen2023teaching}.
\par
\begin{figure}[ht]
    \centering
    \includegraphics[width=1\linewidth]{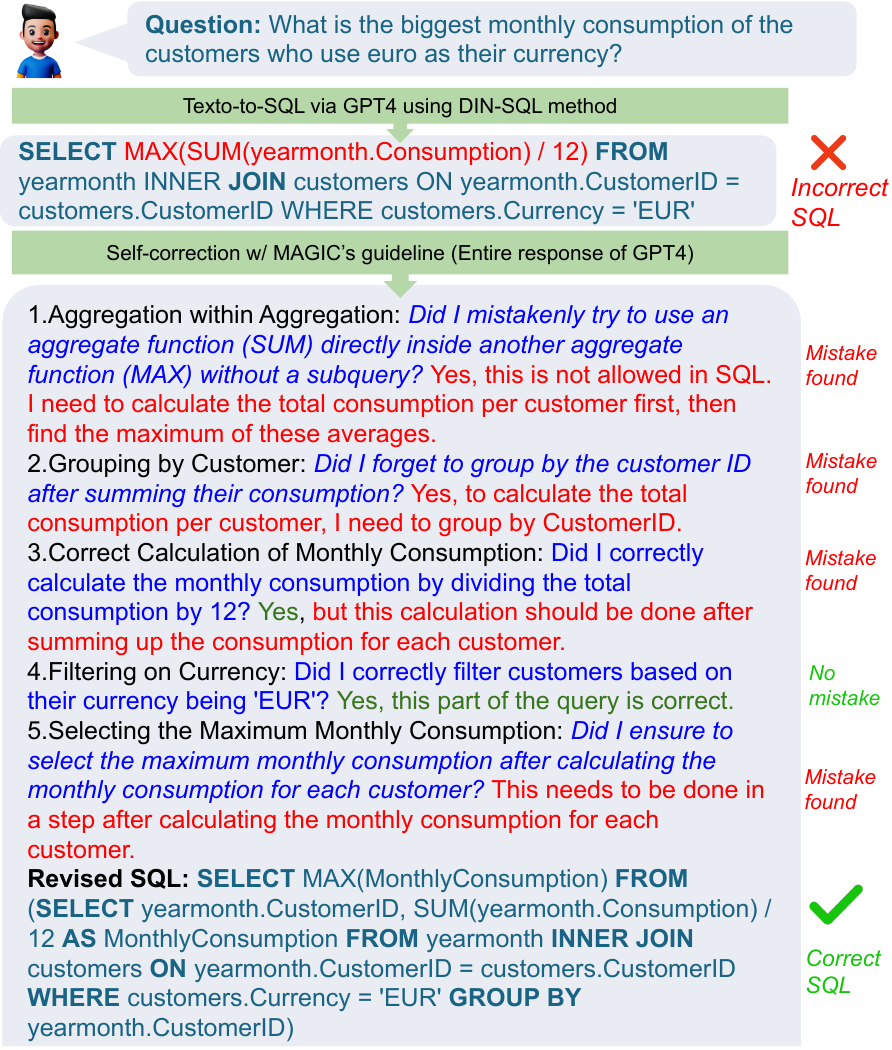}
    \caption{Example of self-correction using automatically generated guidelines by MAGIC}
    \label{fig:example-of-selfcorrection}
\end{figure}
While self-correction for LLMs has been widely studied in variety of tasks including code generation with GPT4 \cite{shypula2024learning,pan2023automatically,kamoi2024can}, it has been relatively unexplored in text-to-SQL.
The common method for self-correcting in existing few-shot LLM based methods that achieve state-of-the-art effectiveness is designing a self-correction guideline by human and prompting it to GPT-4 together with its initial generated SQL \cite{dinsql}.
This self-correction guideline commonly engineered based on the train set where an expert human analyzes common mistakes of the LLM on the training data and design a guideline to prevent common mistakes.
This process is a time-consuming and challenging task that is limited to the ability of humans in identifying all the mistake patterns that exist in the LLM responses \cite{dinsql,macsql,chess}. 
\par
We propose MAGIC, a novel multi-agent self-correction guideline generation for text-to-SQL, that generates an effective self-correction guideline that outperforms human-written guidelines and yields to improving effectiveness of strong few-shot LLM based text-to-SQL methods. 
Similar to humans who engineer the self-correction guideline and apply it to LLMs during inference, with MAGIC, we first tackle guideline generation and then utilize the generated guideline during inference.
By iterating over the incorrect generated SQLs by the initial text-to-SQL method, MAGIC automatically generates the self-correction guideline that is tailored to the mistakes of the initial system.
Next, in inference, the self-correction guideline of MAGIC will be integrated to the initial text-to-SQL method, assisting it in preventing its common mistakes.
\par
Unlike previous studies that analyze self-correction on a simple method that uses GPT4 as its backbone or a low-effective open-source language model \cite{zhang2024benchmarking}, we employ a strong and effective open-source method based on GPT4, DIN-SQL \cite{dinsql}, as our initial text-to-SQL system aiming to compare our automatically generated guideline with an effective expert human-written guideline.
Our experiments demonstrate that the generated guideline by MAGIC leads to self-explanatory self-correction, where the LLM asks itself questions before self-correcting. 
Figure \ref{fig:example-of-selfcorrection} illustrates an example of where the MAGIC guideline assisted GPT4 to self-correct its previously incorrectly generated SQL. The process is interpretable as the LLM first starts with asking itself and then perform self-correction.
\begin{figure*}
    \centering
    \scalebox{0.70}{\includegraphics[]{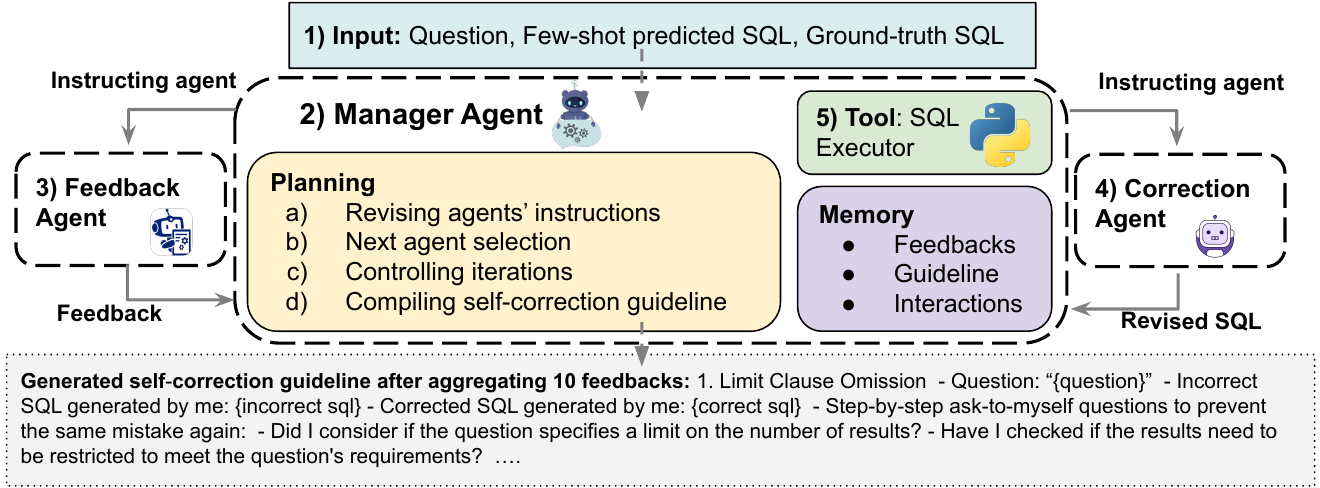}}%
    \caption{Illustration of our proposed method, MAGIC. }
    \label{fig:magic-figure}
\end{figure*}

Our contributions are as follows:
\begin{enumerate*}[label=(\roman*)]
    \item We establish the task of self-correction guideline generation for text-to-SQL, and introduce MAGIC, a novel multi-agent method, that automates the generation of self-correction guideline and outperforms human-written guideline; yielding to improve effectiveness of a strong few-shot LLM based text-to-SQL method. 
    \item We systematically analyze the impact of ours and existing self-correction methods to perform self-correction across different scenarios: Correcting incorrect queries; non-executable queries; and all queries. 
    \item We reproduce all the baselines utilizing the same version of GPT-4 across all the experiments, and provide a comprehensive comparable insight on self-correction in text-to-SQL.
    \item We publish all code to reproduce our experiments as open source.\footnote{\url{https://github.com/microsoft/SynQo}}
\end{enumerate*}
\par
We found that there can be a significant gap in terms of effectiveness when utilizing different versions of GPT-4. For instance, our replication of DIN-SQL \cite{dinsql} on development set of BIRD \cite{li2023can} achieves 56.52 compared to the original report of 50.72 in the paper, which is about 6 points higher than the reported number. This is our motivation for replicating and reproducing all the reported baselines using same GPT-4 and avoid copying numbers from previous papers.  
\section{Related work}
\header{LLMs for Text-to-SQL}
Converting natural language questions into SQL queries, known as text-to-SQL, has been an active area of research within both the natural language processing (NLP) and database communities for many years  \cite{zelle1996learning,guo2019towards}. Recently, text-to-SQL has benefited from the promising effectiveness of Large Language Models (LLMs) \cite{chess,dinsql}. Early methods utilized the zero-shot in-context learning capabilities of LLMs for SQL generation \cite{rajkumar2022evaluating}. Building on this, subsequent models have improved LLM performance through task decomposition and techniques such as Chain-of-Thought (CoT) \cite{wei2022chain}, including but not limited to models like DAIL-SQL \cite{gao2023text}, MAC-SQL \cite{macsql}, C3 \cite{dong2023c3}, self-consistency \cite{wang2022self}, and least-to-most prompting \cite{zhou2022least}. In addition, there are studies focusing on fine-tuning LLMs for text-to-SQL \cite{li2024can,gao2023text,li2024codes}. We choose DIN-SQL \cite{dinsql} as the initial text-to-SQL system due to two reasons: (i) high effectiveness and available open-source code; (ii) its self-correction component is a human expert written guideline based on prior mistakes of the LLM making it a suitable baseline for comparing the automatically self-correction guideline generated by MAGIC with expert human-written guideline. However, we would emphasize that our method is independent of how the in-context learning method is designed and is adaptable to the output of any method, as we only require the predicted SQL of the initial text-to-SQL system for tackling self-correction in MAGIC.
\par
\header{Self-correction in text-to-SQL}
\citet{pan-etal-2024-automatically} and \cite{kamoi2024can} provide an extensive overview of research on self-correction in a variety of domains up to 2024. Here, we focus on self-correction in Text-to-SQL, where there has been relatively limited study. Self-debugging \cite{chen2023teaching} generates additional explanations on the question and initial predicted SQL, which are then provided to an LLM for self-correction. DIN-SQL \cite{dinsql} designs a human-written self-correction guideline and revises all the initially generated SQLs according to this guideline. Self-consistency is based on generating several candidates and employing a voting process, which has been integrated into Text-to-SQL by DAIL-SQL \cite{gao2023text}. The refiner component of MAC-SQL \cite{macsql} uses execution errors as signals for revising the SQL. To the best of our knowledge, we are the first study to address self-correction in Text-to-SQL by generating a self-correction guideline.
\par
\header{LLM-based Agents}
LLM-based agents have been a promising area of study in both academic and industry communities for an extended period \cite{wang2024survey}, leading to extensive research exploring autonomous agents based on LLMs such as AutoGPT \cite{Significant_Gravitas_AutoGPT}, OpenAgents \cite{xie2023openagents}, and AutoGen \cite{wu2023autogen}. However, there are limited studies leveraging this concept for text-to-SQL, with MAC-SQL being the only multi-agent method focusing on addressing the Text-to-SQL task through a new multi-agent collaborative framework \cite{macsql}.
Inspired by the literature on multi-agent LLMs, in MAGIC, we employ a multi-agent collaborative framework that iteratively analyzes the failures of predicted SQLs by a text-to-SQL method and automatically generates a self-correction guideline tailored to the method's mistakes. To the best of our knowledge, there is no prior work in the literature that employs multi-agent methods for designing a self-correction guideline in text-to-SQL or other domains.
\section{Self-correction Guideline Generation}
\header{Tasks definition} 
Given a set of questions in natural language alongside the corresponding database schema and the initially generated incorrect SQL queries by a text-to-SQL system, the task of self-correction guideline generation is to generate a self-correction guideline that is tailored to the LLM's mistakes and prevents it from repeating them. This can be done by a human expert, an LLM, or a collaboration between a human expert and an LLM. The typical approach for self-correction involves writing a guideline for the Large Language Model manually by humans based on the LLM's common mistakes \cite{dinsql,chess}. The data used for guideline generation must not be shared with the evaluation data to prevent overfitting the guidelines on the test errors.
\subsection{MAGIC}
We tackle the task of generating self-correction guideline for text-to-SQL systems, focusing on the failures of an initial method, denoted as $M$, using the training dataset. A failure for $M$ occurs when the predicted SQL is either not executable or its execution result differs from that of the ground truth SQL, denoted as $s^{gt}$. In this context, the predicted SQL that is either not executable or has a differing execution result is called an incorrect SQL, denoted as $s^\prime$.

Our method, MAGIC, consists of three agents illustrated in Figure \ref{fig:magic-figure}: the manager, feedback, and correction agents. These agents interact iteratively over the failures to generate the self-correction guideline. We describe this process in detail in the following. Please note that due to space limitations, the paper includes summarized prompts. We provide the complete prompts and the full automatically generated self-correction guideline by MAGIC in the technical appendix.

\header{Feedback-correction cycle}
Given each question, $s^{gt}$, and $s^\prime$, the manager agent starts a feedback-correction iteration cycle. At each iteration of this cycle, the manager requests the feedback agent for an explanation of the mistakes in $s^\prime$ by comparing it to $s^{gt}$ (steps 2 to 3 in Figure \ref{fig:magic-figure}).
Next, the manager agent integrates the feedback received from the feedback agent to interact with the correction agent. The manager requests the correction agent to revise $s^\prime$ according to the provided feedback (step 3 to 4 in Figure \ref{fig:magic-figure}). Subsequently, the correction agent generates a new revised SQL. At this stage, the manager identifies whether the revised SQL by the correction agent successfully leads to identical results with $s^{gt}$.  To do so, the manager uses its SQL executor tool to execute the correction agent's revised SQL (step 5 in Figure \ref{fig:magic-figure}). 
The manager ends this cycle if the stopping criteria are met. The stopping criteria are either revising $s^\prime$ successfully by the correction agent or reaching the maximum number of iterations. We set $5$ as maximum number of iteration in our experiments.

\header{Revising agents' instruction}
In the first iteration of the feedback-correction cycle, the manager uses two predefined prompts designed for its interactions with the feedback and correction agents, as illustrated in Figures \ref{fig:feedback_prompt} and \ref{fig:correction_prompt} respectively. If the correction agent cannot successfully revise $s^\prime$ in the first iteration, the manager begins revising these predefined prompts. 
The prompt template of manager for revising the predefined prompts of feedback and correction agents are illustrated in Figure \ref{fig:manager_revise_prompt_for_feedback_agent}.
We found empirically that this step plays an important role for the manager to adapt its interaction with agents. Manager revise the agent's instruction based on the previous response of the agent, previous prompt that is used for interacting with the agent, question, $s^\prime$, and $s^{gt}$.

\header{Guideline generation}
Given each successful revision by the correction agent, the manager stores the corresponding successful feedback that led to this in its memory. The manager then aggregates these stored feedbacks to generate the self-correction guideline batch-by-batch using a predefined prompt for guideline generation (Figure \ref{fig:guideline_generation_prompt}). Each batch consist of $k$ feedbacks. In our preliminary experiments, we determined that a feedback batch size of 10 is optimal. In the first batch, the manager begins without any pre-existing guideline and generate an initial guideline. For subsequent batches, the manager updates the already generated guideline. 
This guideline generation process is triggered at the end of a feedback-correction cycle if the manager has accumulated a new batch of successful feedbacks that has not yet been utilized for guideline generation. The self-correction guideline generated by our proposed method, MAGIC, automatically and from scratch, is presented in the technical appendix.%
\begin{figure}[]
    \begin{mdframed}[backgroundcolor=verylightgray,roundcorner=10pt]
    \small 
"question": "\{question\}",\newline
"Correct SQL": "\{Correct SQL\}",\newline
"Incorrect SQL": "\{Incorrect SQL\}". The mistakes are:
    \end{mdframed}
    \centering
    \caption{The predefined prompt for interaction of manager with feedback agent.}
    \label{fig:feedback_prompt}
\end{figure}

\begin{figure}[]
\small 
    \begin{mdframed}[backgroundcolor=verylightgray,roundcorner=10pt]
- Schema Overview:  \{schema\} \newline
- Question: \{question\}  \newline
- Predicted SQL: ```sql \{Incorrect SQL\} ```\newline
- Expert Human Feedback: \{feedback\}
    \end{mdframed}
    \centering
    \caption{The predefined prompt for interaction of manager with correction agent.}
    \label{fig:correction_prompt}
\end{figure}

\begin{figure}[]
    \begin{mdframed}[backgroundcolor=verylightgray,roundcorner=10pt]
    \small 
Manager, review the following prompt for the agent and generate a revised prompt. \newline
Agent description: \{Agent description\}.\newline
Previous output and prompt that was not useful: \{agent\_outputs[-1]\} and \{agent\_prompts[-1]\}\newline
Revise prompt (Return the entire prompt):
    \end{mdframed}
    \centering
    \caption{The prompt template of manager for revising the predefined prompts of feedback or correction agent.}
    \label{fig:manager_revise_prompt_for_feedback_agent} %
\end{figure}

\begin{figure}[]
    \begin{mdframed}[backgroundcolor=verylightgray,roundcorner=10pt]
    \small 
\# Recent mistakes to add to Guideline:\newline
\{batch\_of\_successful\_feedbacks\}\newline
\# Updated Guideline (Return the entire guideline):
    \end{mdframed}
    \centering
    \caption{The prompt of manager for self-correction guideline generation.}
    \label{fig:guideline_generation_prompt}
\end{figure}
\header{Efficiency}
Previous approaches to craft self-correction guidelines needed extensive expert human work. With MAGIC, we can efficiently generate, empirically shown better guideline, in less than 2 hours. 

\subsection{Agents access to information}
Although the manager agent has access to all the available information during generation of self-correction guideline, the agents has limited access to the information about the task. 
During the iterations, the correction agent consistently receives only $s^\prime$.
This ensures that a successful self-correction can indicate that the feedback could effectively cover the mistakes that exist in the $s^\prime$. This is important since the goal is generating a guideline that can prevent from initial text-to-SQL system errors. As the result, the self-correction agent should focus on correcting the initial system prediction.

\begin{table*}[]
\small
\centering
\scalebox{0.755}{
 \begin{tabular}{p{4cm}|cccc|cccc|cccc}
 \toprule
 & \multicolumn{12}{|c}{Scenario of Self-Correction Application} \\ \midrule
 \multirow{2}{*}{DIN-SQL}& \multicolumn{4}{c|}{Correcting incorrect SQLs} & \multicolumn{4}{c|}{SQLs with Exec error} & \multicolumn{4}{c}{All SQLs} \\ \cmidrule(lr){2-5} \cmidrule(lr){6-9} \cmidrule(lr){10-13} 
 \ & S & M& C & T & S & M & C & T & S & M & C & T \\ \midrule 
W/o self-correction & 63.14 & 49.03 & 38.19 & 56.52 & 63.14 & \underline{49.03} & 38.19 & 56.52 & 63.14 & 49.03 & 38.19 & 56.52 \\ \midrule 
\multicolumn{13}{c}{Self-correction w/o guideline} \\ \midrule %
Self-Debugging \cite{chen2023teaching} & 63.57 & 50.11 & 39.67 & 57.24& 63.35 & \underline{49.03}  & 39.58 & 56.78 & 63.03 & 49.25 & 38.89 & 56.58\\ 
Self-Consistency \cite{wang2022self} & \underline{64.56} & \underline{50.79} & \underline{40.09}& \underline{58.08} &  63.24 & \underline{49.03}  & 38.19  & 56.58 & 64.00 & \underline{49.68} & 37.50 & 57.17\\
Multiple-Prompt \cite{lee2024mcs} & 64.22 & 50.75 & 40.02& 57.86 & 63.35 & \underline{49.03}  & 38.19  & 56.65 & 63.68 & \textbf{50.11} & 39.58 & 57.30 \\ \midrule 
\multicolumn{13}{c}{Self-correction w/ guideline} \\ \midrule 
Human Expert G, MAC-SQL \cite{macsql} & 63.18 & 49.19 & 38.30& 56.60 & 63.20 & \underline{49.03}  & 38.19  & 56.58 & 47.80 & 46.22 & 35.30& 46.14 \\
Human Expert G, DIN-SQL  \cite{dinsql} &  64.32 &  50.32 & 39.58 & 57.76  & \underline{63.58}  & \underline{49.03} & \underline{38.89} & \underline{56.98} & 62.49 & 48.17 & 37.50 & 55.80 \\ %
MAGIC G \textbf{(ours)} &  \textbf{65.84$^\dagger$} & \textbf{51.61$^\dagger$} & \textbf{40.28$^\dagger$} & \textbf{59.13$^\dagger$}& \textbf{63.69$^\dagger$}&\textbf{50.15$^\dagger$} & \textbf{39.59$^\dagger$} &\textbf{57.32$^\dagger$} & \textbf{65.75$^\dagger$} & 49.46 & \textbf{41.67$^\dagger$} & \textbf{58.55$^\dagger$}\\ 
 \bottomrule
 \end{tabular}
}
\caption{Effectiveness results in terms of Execution Accuracy (EX) on the DEV set of the BIRD dataset. All the experiments have been reproduced by us. `S', `M', `C', and `T' refer to different levels of difficulty: Simple, Medium, Challenging, and Total, respectively. The baseline for Natural Language to SQL conversion is DIN-SQL and `Human Expert G' refers to self-correction guideline (prompt). `MAGIC refers to the guideline that is automatically generated by our proposed method, MAGIC, using the train set of BIRD dataset. Significance is shown with $\dagger$ for MAGIC compared to both `Human Expert' G baseliens. Statistical significance was measured with a paired t-test ($p<0.05$) with Bonferroni correction for multiple testing.}

\label{tab:guideline_res}
\end{table*}

\section{Experimental setup}
\header{Datasets}
The Spider \cite{yu-etal-2018-spider} dataset is a collection of 10,181 questions and 5,693 unique complex SQL queries across 200 databases in 138 domains, with each domain featuring multiple tables. It is divided into training, development, and test sets with 8,659, 1,034, and 2,147 examples, respectively, across 146, 20, and 34 distinct databases, ensuring no overlap between sets. Queries are classified into four difficulty levels based on complexity factors such as SQL keywords, nested subqueries, and use of column selections and aggregations.
The BIRD dataset \cite{li2023can} comprises 12,751 unique question-SQL pairings across 95 relative large databases (33.4 GB) in 37 professional domains like blockchain and healthcare. BIRD introduces external knowledge as an additional resource for generating accurate SQL queries to bring more complexity into the task.
\par
\header{Metrics}
We employ two key metrics from existing literature: Execution Accuracy (EX) and Valid Efficiency Score (VES). Execution Accuracy measures the correctness of a predicted SQL query by comparing its execution output with that of the ground truth SQL query. A SQL query is deemed correct if its execution results match those of the ground truth, allowing for a precise assessment of the model's performance given the possibility of multiple valid SQL queries for a single question. The Valid Efficiency Score evaluates the efficiency of executing the generated SQL queries, focusing on both their accuracy and execution time, but only considers queries that produce correct results, i.e., those whose execution outcomes are consistent with the reference query. This dual-metric approach provides a comprehensive evaluation of the model's ability to generate both accurate and efficient SQL queries. Given our primary focus on enhancing accuracy, we prioritize Execution Accuracy (EX) as the main metric in our experiments, while also reporting the Valid Efficiency Score (VES) in selected cases to offer additional insights.  
Furthermore, we do not report VES results for the SPIDER dataset because computing VES for this dataset is less meaningful. The execution times for SPIDER are so short that any time differences often round to zero, rendering VES comparisons inconclusive. This is why previous studies also omit VES for SPIDER dataset \cite{dinsql,chess}. The SPIDER dataset's databases contain significantly fewer entries compared to the BIRD dataset, where the databases are, on average, 140 times larger. This substantial difference in scale makes VES a more relevant metric for BIRD, where efficiency considerations are more significant.
\par
\textbf{Baselines.} 
We reproduce DIN-SQL using the publicly available original implementation \cite{dinsql}. For self-correction, we employ an extensive set of baselines categorized into guideline-dependent and guideline-independent methods. For guideline-independent methods, we reproduce self-debugging \cite{chen2023teaching} based on the written prompts in the paper. For self-consistency \cite{wang2022self,gao2023text}, we generate 20 SQL queries given the SQL generation prompt of DIN-SQL instead of generating only one SQL and select the final SQL based on voting. In voting, we execute all SQL queries and select the most frequently returned result's SQL. If one result is produced by various SQL queries, we select the most efficient one by comparing them against each other. For the Multiple-Prompt baseline, we follow the approach in \cite{lee2024mcs} by re-ordering candidate tables in the prompt and generating up to 20 different combinations, employing a voting mechanism similar to our self-consistency implementation. For human-expert baselines, we utilize the self-correction guideline from DIN-SQL \cite{dinsql}, written by human experts, and the revision guideline from MAC-SQL \cite{macsql}, that is also written by human experts and are designed specifically for correcting SQL queries with execution errors. However, we also adopt their approach for correcting incorrect SQL queries and for correcting all initially predicted SQL queries, regardless of whether they are correct, similar to the scenarios analyzed in \cite{chen2023teaching} and \cite{dinsql} respectively. %
Our method, MAGIC, is designed to generate self-correction guidelines for text-to-SQL tasks without relying on fine-tuned models. We exclude baselines that depend on fine-tuning \cite{chess,pourreza2024dts,li2024dawn} to ensure fair comparisons and to highlight MAGIC's effectiveness in scenarios where fine-tuning is not feasible due to high computational costs.

\section{Results}
This section addresses the following research questions (RQs):
\begin{itemize}[leftmargin=*]
\item \textbf{RQ1:} What is the effectiveness of MAGIC's self-correction guideline compared to the existing state-of-the-art baselines for self-correction in text-to-SQL, particularly across different error scenarios and datasets? %
\item \textbf{RQ2:} How is the effectiveness of generated guidelines influenced by the quantity of feedback - is more feedback always better?
\item \textbf{RQ3:} What is the impact of manager agent intervention on reducing the number of iterations and increasing the number of corrected SQLs? 
\end{itemize}
\header{Main results (RQ1)}
Table \ref{tab:guideline_res} and \ref{tab:spider_res} present the results of self-correction with the guidelines generated by MAGIC on the development set of BIRD and SPIDER, respectively. 
Overall, the guidelines generated by our proposed method outperform all the baselines in terms of total effectiveness measured by execution accuracy and demonstrate significant improvements in self-correction performance across the BIRD and SPIDER datasets, outperforming self-correction guidelines written by expert humans. For example, MAGIC's guidelines improve the DIN-SQL \cite{dinsql} baseline from 56.52 to 59.13 in terms of execution accuracy and outperform the self-correction guideline of \cite{macsql} as well, as presented in Tables \ref{tab:different_methods_res} and \ref{tab:guideline_res} respectively.

We experiment with self-correction in three scenarios, as shown in Table \ref{tab:guideline_res}: (i) correcting incorrect SQLs, assuming the availability of a correctness oracle. This setup is common in recent studies focusing on self-correction in text-to-SQL \cite{chen2023teaching, zhang2024benchmarking}. (ii) SQLs with execution errors: this scenario is common in self-correction where there is no available correctness oracle, but the goal is to avoid generating incorrect and non-executable SQL \cite{macsql}. (iii) All SQLs: this setup is common in methods where self-correction is part of a multi-step model, employing a self-correction guideline on all predictions \cite{dinsql}.

\begin{table}[]
\centering
\scalebox{0.85}{
\begin{tabular}{llc}
\toprule
\multicolumn{2}{l}{Correction Method}&  EX \\ \midrule  
& w/o self-correction  & 78.62 \\ \midrule 
\multicolumn{3}{l}{Self-correction w/o guideline} \\ 
& Self-Consistency \cite{wang2022self}& \underline{81.64}  \\
& Multiple-Prompt \cite{lee2024mcs} & 80.22  \\  \midrule
\multicolumn{3}{l}{Self-correction w/ guideline} \\ 
& Human's G \cite{macsql} & 81.15  \\
& Human's G \cite{dinsql} &  80.35  \\
& MAGIC's G (Ours) &  \textbf{85.66} \\
\bottomrule
\end{tabular}
}

\caption{Effectiveness results for self-correcting incorrect SQL queries on the SPIDER development set. We compared MAGIC's generated guideline (MAGIC's G) against two top self-correction baselines: self-consistency and multiple-prompt, as well as expert human self-correction guidelines.}%
\label{tab:spider_res}
\end{table}

\begin{table}[]
\centering
\scalebox{0.9}{
\begin{tabular}{l|ccccc}
\toprule
& \multicolumn{5}{c}{\# of aggregated batch of feedbacks} \\  \cmidrule(lr){2-6}
& 0 & 1 & 5 & 10 & 39 (All) \\ \midrule
EX & 56.52 & 57.4 & 58.8 & 59.13 & 59.13 \\
\bottomrule
\end{tabular}
}
\caption{Impact of number of aggregation levels of feedbacks on the effectiveness of generated self-correction guideline on the development set of BIRD dataset.}%
\label{tab:feedback_aggregation_effect}
\end{table}

We find that applying self-correction to all predicted SQLs, as in the human-expert written guideline for self-correction in \cite{macsql}, reduces effectiveness. Intuitively, when self-correction is applied for all predicted SQLs, there is a risk that the LLM changes its previously correct prediction to an incorrect one, as suggested by previous work \cite{Confidence2024}. Therefore, designing a proper self-correction guideline is crucial, or it is safer to apply self-correction where a correctness oracle can identify whether the initial response is correct or not. This oracle can be a human user who minimally interacts with the system and only determines whether the executed query results meet their expectations. Regarding the "SQLs with execution errors" scenario, only 46 queries predicted by the initial text-to-SQL system \cite{dinsql} are non-executable, which limits the effectiveness of self-correction for these cases, with several methods achieving the same effectiveness (49.03) for medium query difficulty. It is noteworthy that although we have not attempted to combine MAGIC with other baselines, as our focus is on analyzing different self-correction methods rather than integrating them, some baselines could potentially enhance MAGIC if combined, such as Self-consistency \cite{wang2022self} and Multiple-Prompt \cite{lee2024mcs}.%
\begin{figure}[]
    \centering
    \begin{minipage}{0.45\textwidth}
        \centering
        \scalebox{0.80}{\includegraphics[width=\linewidth]{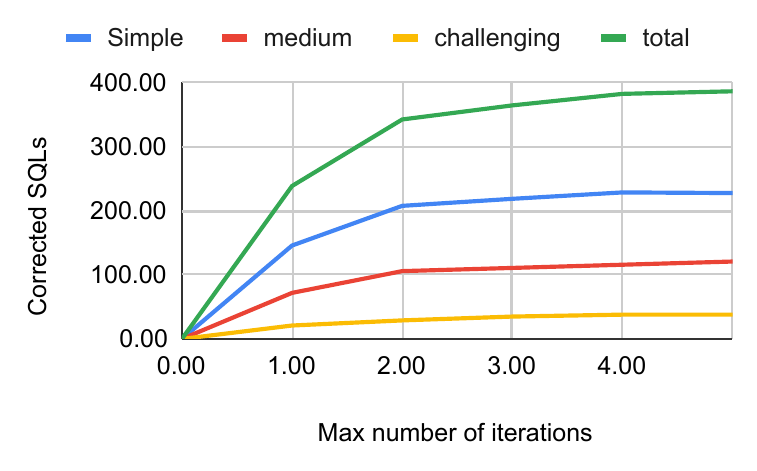}}
        \caption*{a) MAGIC w/ manager.}
        \label{fig:magic_with_manager}
    \end{minipage}\hfill
    \begin{minipage}{0.45\textwidth}
        \centering
        \scalebox{0.80}{\includegraphics[width=\linewidth]{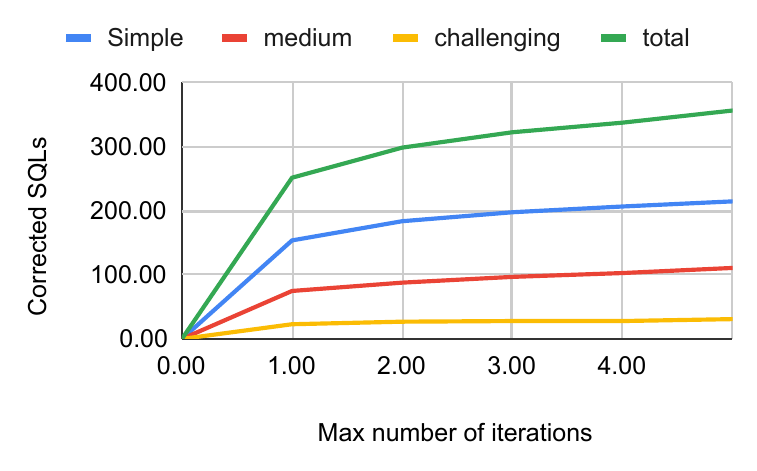}}
        \caption*{b) MAGIC w/o manager.}
        \label{fig:magic_without_manager}
    \end{minipage}
    \caption{Analyzing impact of maximum number of iterations on the total number of corrected SQLs, that were initially incorrect, with and without manager being involved in MAGIC. The analysis is conducted on the train set of the BIRD dataset.}
    \label{fig:iterations-vs-corrected-sqls}
\end{figure}
\par
\header{Impact of Feedback Quantity (RQ2)} Table \ref{tab:feedback_aggregation_effect} presents the results of the generated self-correction guideline effectiveness by MAGIC, where different numbers of batches of feedback are aggregated for generating the guideline. We found that after aggregating 10 batches of feedback, each batch consist of 10 feedbacks, the generated guideline by MAGIC outperforms existing self-correction baselines, including those with guidelines written by humans. This shows the efficiency of process of MAGIC for self-correction guideline generation. Furthermore, we found that questions for which MAGIC cannot provide feedback to self-correct are often controversial. For example, when an additional aggregation function column is reported alongside the main column name that should be reported, such as 'SELECT city, MAX(population) FROM cities` where the MAX(population) should not be included in the predicted query according to the ground truth. 
\begin{table}[]
\centering
\scalebox{0.76}{
\begin{tabular}{l|ccc|cc}
    \toprule
    \multirow{2}{*}{} &\multicolumn{3}{c|}{BIRD} & \multicolumn{2}{c}{Spider} \\ \cmidrule(lr){2-4} \cmidrule(lr){5-6} 
    & EX $\uparrow$ & VES $\uparrow$ & AVG$_i$ $\downarrow$ & EX $\uparrow$ & AVG$_i$ $\downarrow$ \\ \midrule 
    \multicolumn{6}{c}{MAGIC$_{\text{max}_{i}=2}$} \\ \midrule
    w/o manager & 73.01 & 68.29 & 1.17 & 81.40 & 1.11 \\
    w/ manager & \underline{78.94} & \underline{80.22} & \textbf{1.12} & \underline{86.52} & \underline{1.05} \\ \midrule
    \multicolumn{6}{c}{MAGIC$_{\text{max}_{i}=5}$} \\ \midrule
    w/o manager & 78.60 & 79.58 & 2.87 & 83.28 & 2.22 \\ %
    w/ manager & \textbf{81.81} & \textbf{84.19} & \underline{2.41} & \textbf{91.78} & \textbf{2.15} \\ 
    \bottomrule
\end{tabular}
}
\caption{The effectiveness for generating feedbacks on the train set of BIRD dataset. The $max_i$ and $AVG_i$ refers to the maximum number of iterations and average number of iterations till stopping criteria triggers respectively.} 
\label{tab:magic_ablation}
\end{table}

\par
\header{Feedback generation effectiveness (RQ3)} 
Figure \ref{fig:iterations-vs-corrected-sqls} presents an analysis of the number of corrected SQL queries with and without the inclusion of the manager agent, by evaluating the performance of the MAGIC framework across a range of maximum iteration counts used as stopping criteria, from one to five. When the manager agent is excluded, the feedback agent controls the maximum iteration count and utilizes the SQL executor tool to compare the execution results of the revised SQL with the ground truth SQL. This comparison underscores the significant role of the manager agent in correcting a larger number of incorrect SQL queries within fewer iterations and ultimately correcting more SQL queries when both setups reach the maximum of five iterations. It is important to note that excluding the manager agent from MAGIC is equivalent to the critic-refine cycle method \cite{kamoi2024can}, that has been employed in other tasks, such as code generation. However, unlike MAGIC, the critic-refine \cite{kamoi2024can} does not produce any guidelines as a result of its feedback generation process and has not been applied to text-to-SQL tasks to the best of our knowledge.
\par
\begin{table}[]
\small
\centering
\scalebox{.8}{
\begin{tabular}{lc}
    \toprule
     Method & EX\\ \midrule
    Zero-shot GPT-4 \cite{openai2023gpt} & 40.18 \\
    Zero-shot GPT-4 + MAGIC G (Ours)  & 48.19 \\ \midrule
    Few-shot CoT GPT-4 \cite{li2023can} &  45.66 \\
    Few-shot CoT GPT-4 \cite{li2023can} + MAGIC G (Ours)  &50.38 \\ \midrule
    DIN-SQL \cite{dinsql} &  56.52 \\
    DIN-SQL \cite{dinsql} + MAGIC G (Ours)  & 59.13 \\ \midrule
    MAC-SQL \cite{macsql} &  \underline{59.39} \\
    MAC-SQL \cite{macsql} + MAGIC G (Ours)  & \textbf{61.92} \\ 
    \bottomrule
\end{tabular}
}
\caption{The effectiveness of self-correction guideline generated by MAGIC across different methods.}
\label{tab:different_methods_res}
\end{table}
We further analyze the impact of manager agent in terms of overall performance of self-correction considering all the queries. Table \ref{tab:magic_ablation} measures the effectiveness of MAGIC in terms of execution accuracy (EX) to determine how many could be addressed correctly in total with the already corrected predictions by LLM and self-corrected ones with MAGIC. We also measure Valid Efficiency Score (VES) to determine whether the self-correction by MAGIC results in optimized SQLs compared to the ground truth. 
As it can be observed,  MAGIC with the manager agent outperforms the exclusion of manager agent in all setups in terms of all metrics across the BIRD and SPIDER datasets.
Results show that MAGIC with manager agent is able to generate more efficient SQLs compared to exclusion of manager agent, e.g., 84.19 vs. 79.58 in terms of VES for w/ manager and w/o manager on the BIRD dataset where the maximum number of iteration is set to 5. We manually analyzed the revised prompt by manager agent and found out during its interactions with the agents where it revises the prompts, the manager agent tends to add instructions for self-correction to optimize the efficiency of SQL, which might be the reason behind generating more efficient SQLs.
\begin{figure}[]
    \centering
    \scalebox{0.37}{\includegraphics[]{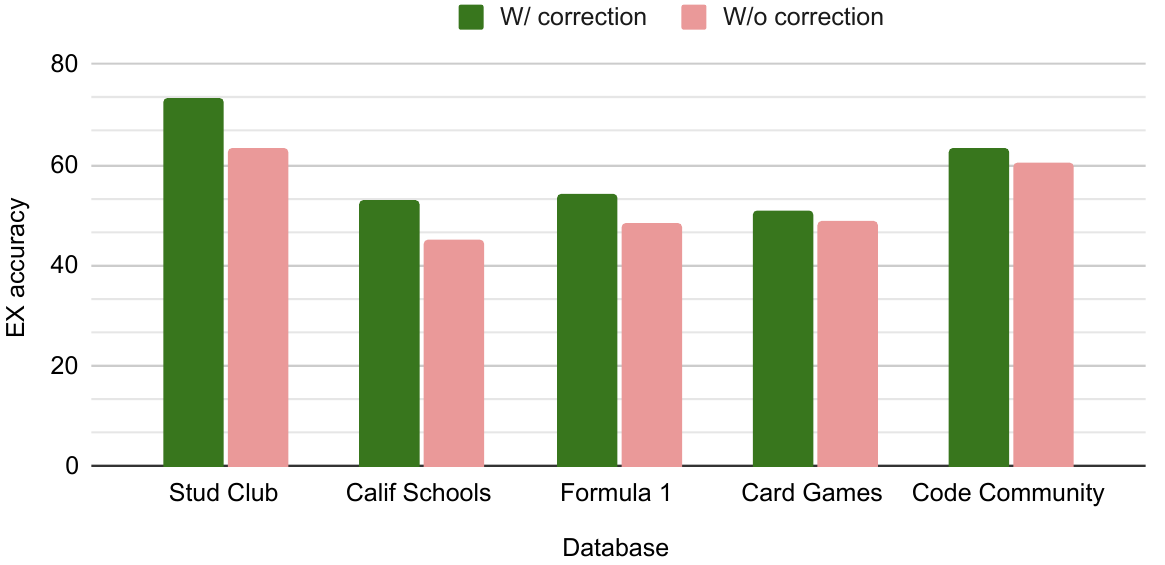}
}
    \caption{Analyzing impact of MAGIC's correction guideline across different databases of BIRD dataset.}
    \label{fig:across-databases}
\end{figure}
\section{Discussion}
\header{Applicability to other methods} We analyze the applicability of the guideline generated by MAGIC based on the mistakes of the DIN-SQL \cite{dinsql} method on three other methods that generate SQL in a single step: zero-shot GPT-4 \cite{openai2023gpt4}, few-shot CoT GPT-4 \cite{li2023can}, and multi-agent (MAC-SQL \cite{macsql}) as shown in Table \ref{tab:different_methods_res}. As we can observe, MAGIC's guidelines improve the zero-shot GPT-4 baseline from 40.18 to 48.19, Few-shot CoT GPT-4 from 45.56 to 50.38, and multi-agent baseline from 59.39 to 61.92 in terms of execution accuracy, respectively.
\par
\header{Source of examples in guideline} 
We analyzed the relationship between the generated guideline and the feedback that contributed to its creation. Specifically, we examined whether any examples in the guideline directly mirrored instances from the feedback or outputs of the correction agent. Our findings indicate that no example was directly copied from the feedback to be used in the guideline. Instead, our empirical observations reveal that the manager agent tends to aggregate multiple pieces of feedback to construct new examples into the self-correction guideline.
\par
\header{Database analysis} Figure \ref{fig:across-databases} illustrates the impact of the self-correction guidelines generated by MAGIC across different databases in the development set of the BIRD dataset. We found that while self-correction improves effectiveness across all databases, some databases benefit more from MAGIC's guidelines. This variability could be due to the capability of the LLM, the differing levels of challenging information in the databases, or other factors. This observation could motivate the development of guidelines that are adapted to difficulty of database in the future.
\section{Conclusions}
This paper presents a new perspective on self-correction in in-context learning for text-to-SQL translation. It proposes a novel method for generating self-correction guidelines, called MAGIC. The motivation behind this approach is to overcome the limitations of existing methods that generate self-correction guidelines by hand, a time-consuming task. Additionally, it addresses the important and costly task of automatically fixing incorrect SQL generated by humans. This work showcases the potential of leveraging LLMs to generate their own self-correction guidelines and highlights the significance of guideline generation in text-to-SQL. We emphasize the importance of improving self-correction methods in text-to-SQL and addressing them as a separate task. The findings of this study contribute to advancing the state-of-the-art in text-to-SQL translation, as our method can be applied to fix issues in any method and provide valuable insights for future research in this domain.

\clearpage
\bibliography{aaai25}

\clearpage
\appendix

\section{Prompts}
\begin{figure*}
    \centering
    \includegraphics[angle=90,origin=c,height=0.67\textheight]{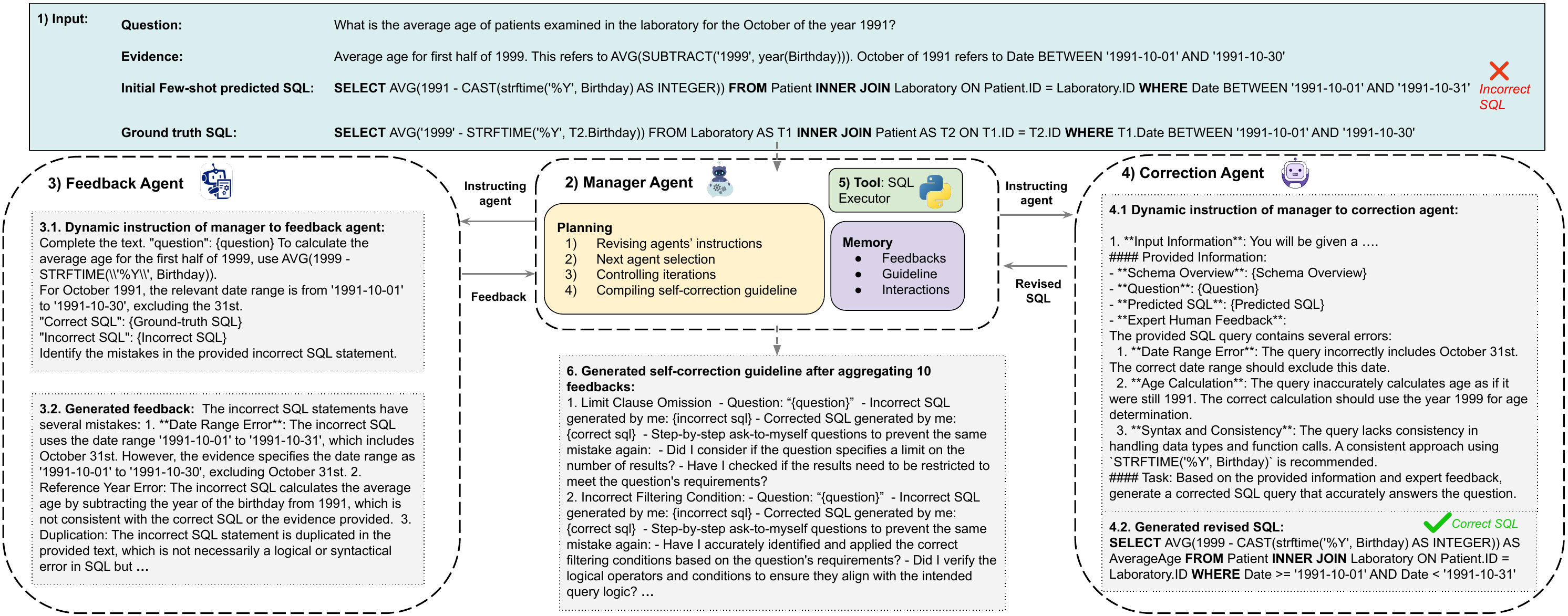}
    \caption{Illustration of last iteration of our proposed method, MAGIC, on a question of BIRD dataset, that led to a successful self-correction.}
    \label{fig:magic-figure-process}
\end{figure*}

The prompts that are used are as follow:
\begin{itemize}
    \item The prompt for the predefined interaction of manager agent with feedback agent can be seen in Figure \ref{fig:feedback_prompt_appendix}, and the prompt with that the manager agent revises its previous predefined prompt can be seen in Figure \ref{fig:manager_revise_prompt_for_feedback_agent_appendix}.
    \item The prompt for the predefined interaction of manager agent with feedback agent can be seen in Figure \ref{fig:correction_prompt_appendix}, and the prompt with that the manager agent revises its previous predefined prompt can be seen in Figure \ref{fig:manager_revise_prompt_for_correction_agent_appendix}.
    \item The prompt for manager agent to generate the self-correction guideline can be seen in Figure \ref{fig:guideline_generation_prompt_appendix}.
\end{itemize}
Furthermore, an example of last iteration of our proposed method, MAGIC, on a question of BIRD dataset, that led to a successful self-correction can be seen in Figure \ref{fig:magic-figure-process}.
\section{Generated Self-correction guideline by MAGIC}
\begin{figure*}[h]
   \centering
   \begin{tabular}{@{}c@{\hspace{.5cm}}c@{}}
       \includegraphics[page=1,width=0.95\textwidth]{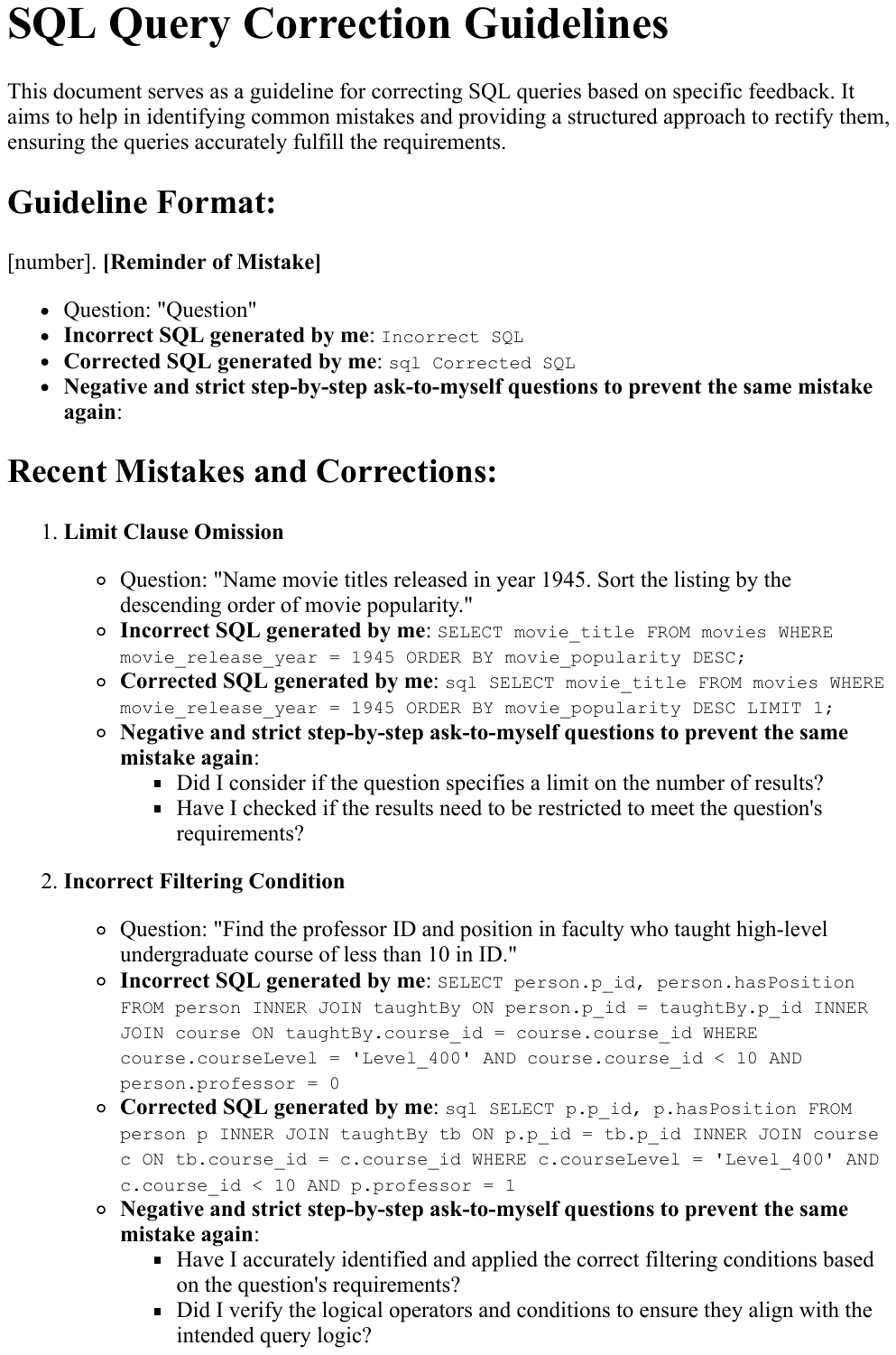} & 
   \end{tabular}
 \caption{Self-correction guideline that is generated by our method, MAGIC, automatically. Page 1.}
 \label{fig:guideline-magic-page1}
\end{figure*}

\begin{figure*}[h]
   \centering
   \begin{tabular}{@{}c@{\hspace{.5cm}}c@{}}
       \includegraphics[page=2,width=0.95\textwidth]{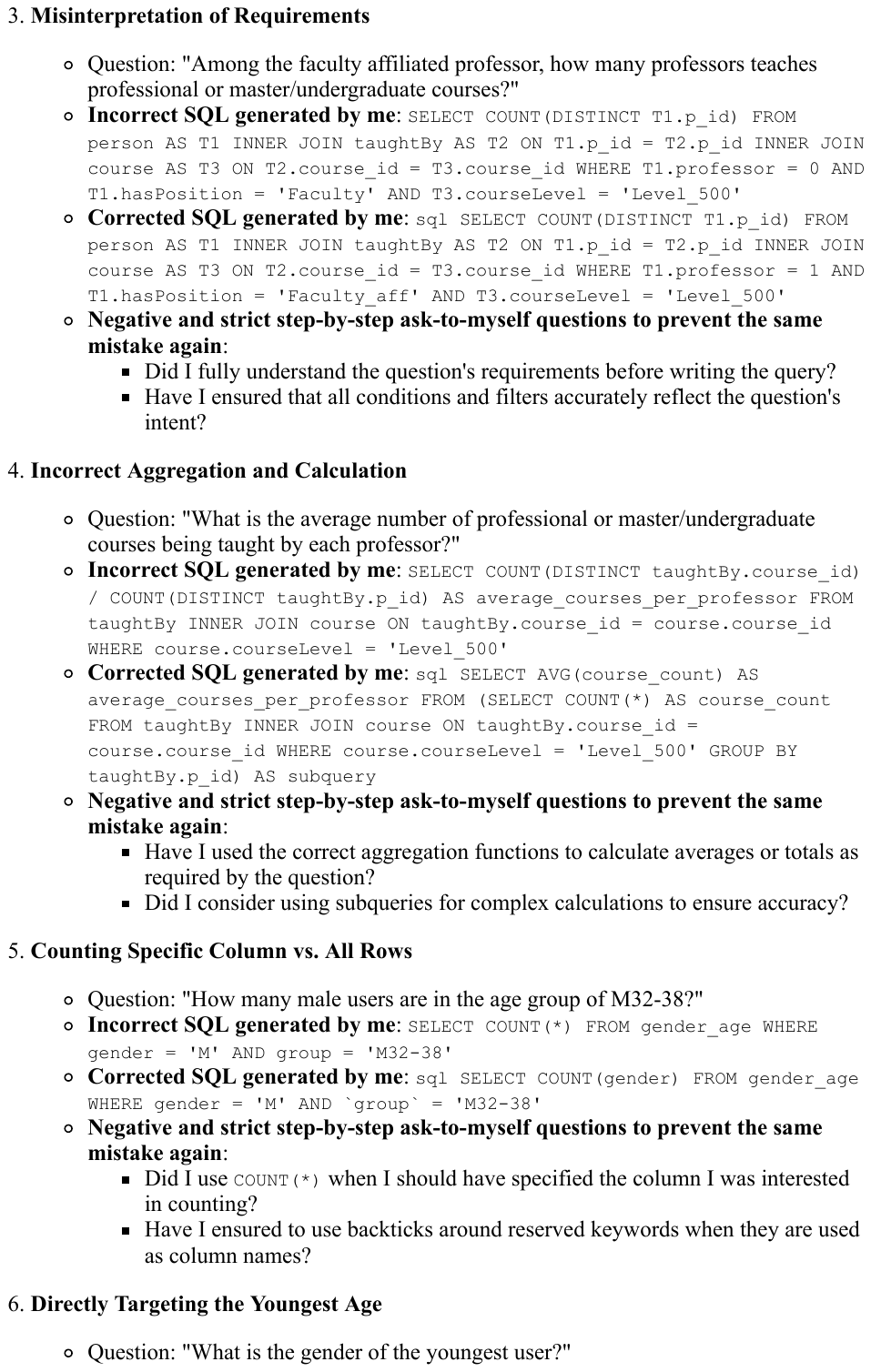} \\[.5cm]
   \end{tabular}
 \caption{Self-correction guideline that is generated by our method, MAGIC, automatically. Page 2.}
 \label{fig:guideline-magic-page2}
\end{figure*}

\begin{figure*}[h]
   \centering
   \begin{tabular}{@{}c@{\hspace{.5cm}}c@{}}
       \includegraphics[page=3,width=0.95\textwidth]{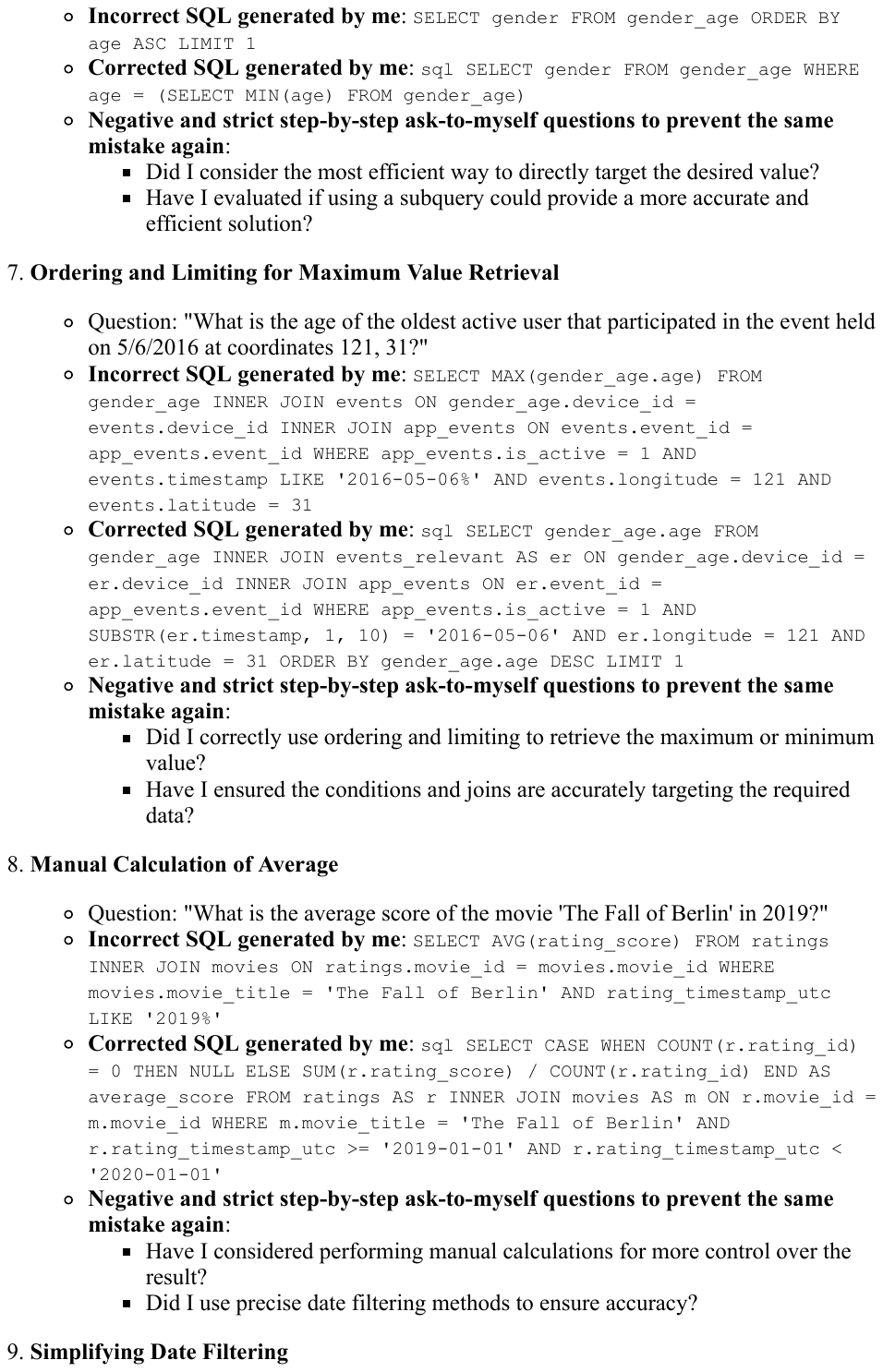} \\[.5cm]
   \end{tabular}
 \caption{Self-correction guideline that is generated by our method, MAGIC, automatically. Page 3.}
 \label{fig:guideline-magic-page3}
\end{figure*}

\begin{figure*}[h]
   \centering
   \begin{tabular}{@{}c@{\hspace{.5cm}}c@{}}
       \includegraphics[page=4,width=0.95\textwidth]{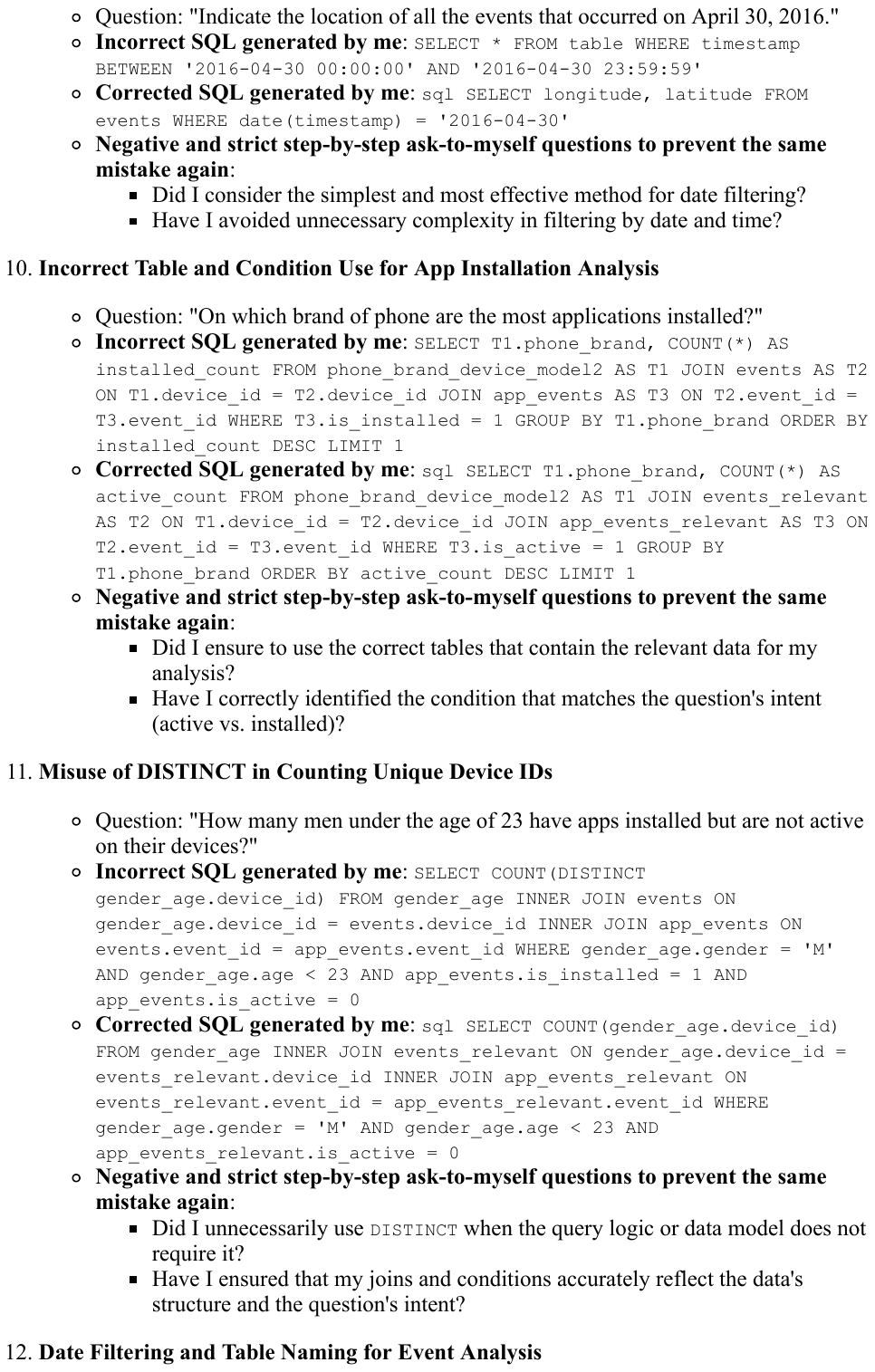} \\[.5cm]
   \end{tabular}
 \caption{Self-correction guideline that is generated by our method, MAGIC, automatically. Page 4.}
 \label{fig:guideline-magic-page4}
\end{figure*}

\begin{figure*}[h]
   \centering
   \begin{tabular}{@{}c@{\hspace{.5cm}}c@{}}
       \includegraphics[page=5,width=0.95\textwidth]{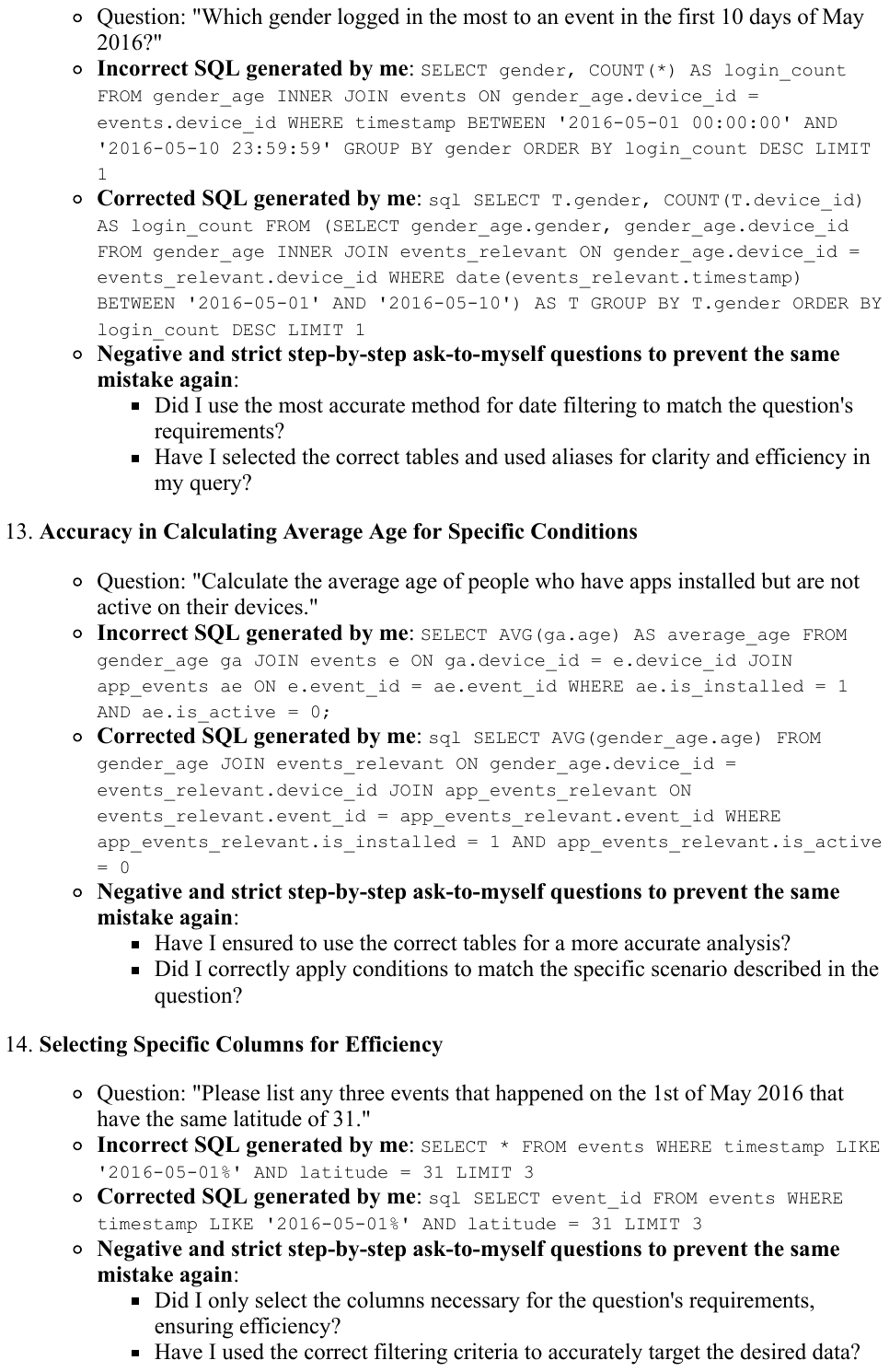} \\[.5cm]
   \end{tabular}
 \caption{Self-correction guideline that is generated by our method, MAGIC, automatically. Page 5.}
 \label{fig:guideline-magic-page5}
\end{figure*}

\begin{figure*}[h]
   \centering
   \begin{tabular}{@{}c@{\hspace{.5cm}}c@{}}
       \includegraphics[page=6,width=0.95\textwidth]{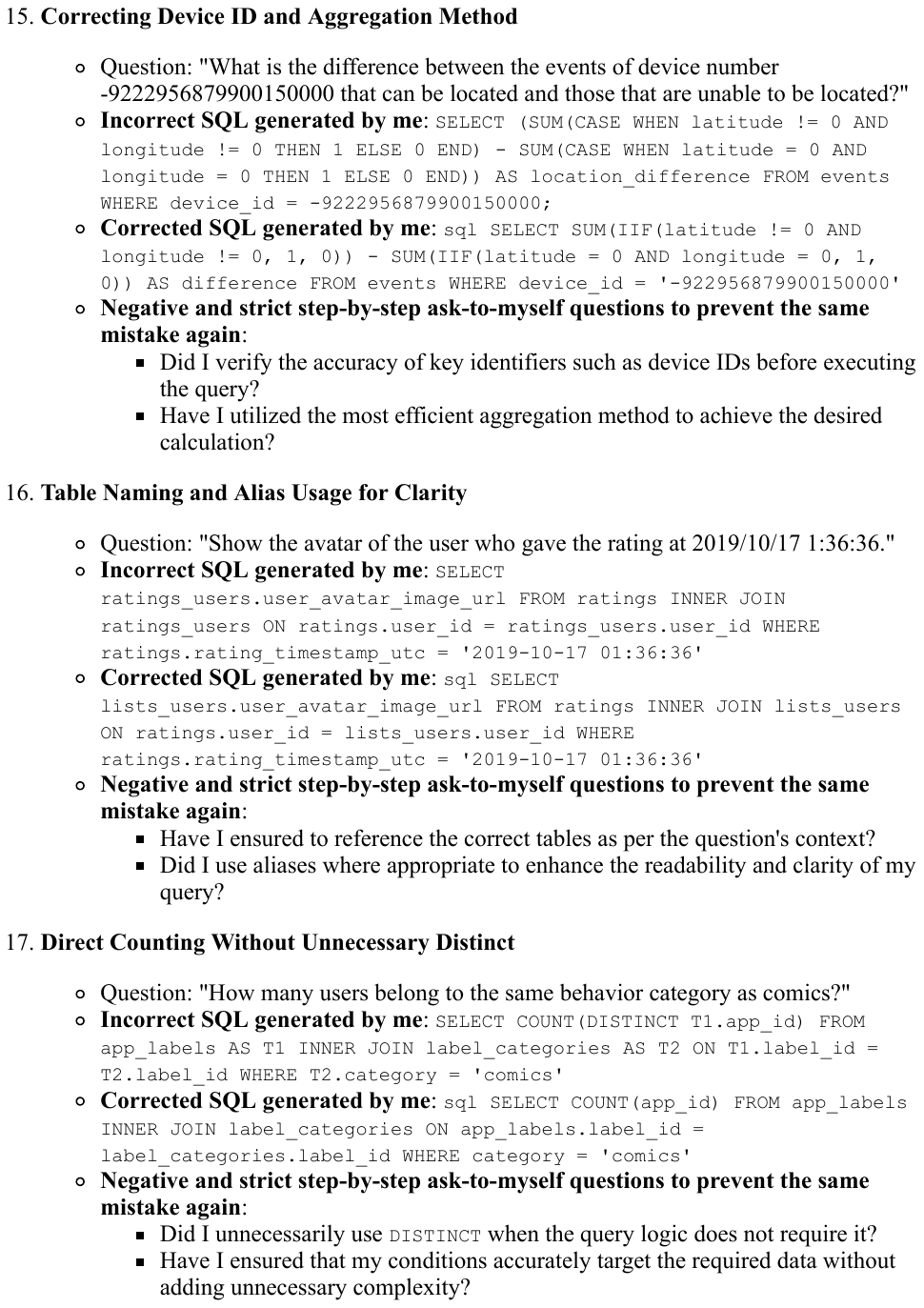} \\[.5cm]
   \end{tabular}
 \caption{Self-correction guideline that is generated by our method, MAGIC, automatically. Page 6.}
 \label{fig:guideline-magic-page6}
\end{figure*}

The self-correction guideline generated by our proposed method, MAGIC, automatically and from scratch, is presented in Figures \ref{fig:guideline-magic-page1}, \ref{fig:guideline-magic-page2}, \ref{fig:guideline-magic-page3}, \ref{fig:guideline-magic-page4}, \ref{fig:guideline-magic-page5}, and \ref{fig:guideline-magic-page6}. 
 
\begin{figure}[ht]
    \begin{mdframed}[backgroundcolor=verylightgray,roundcorner=10pt]
\textbf{System role prompt:} \newline \newline
Complete the text in chat style like a database manager expert. Write in simple present without using correct SQL. Accept what the user identifies as correct or incorrect.
\newline \newline
\textbf{User role prompt:} \newline \newline
"question": "\{question\}",\newline
"evidence": "\{evidence\}",\newline
"Correct SQL": "\{Correct SQL\}",\newline
"Incorrect SQL": "\{Incorrect SQL\}",\newline \newline
Incorrect SQL mistakes are:
    \end{mdframed}
    \centering
    \caption{The predefined prompt for interaction of magic agent with feedback agent.}
    \label{fig:feedback_prompt_appendix}
\end{figure}
\begin{figure}[]
    \begin{mdframed}[backgroundcolor=verylightgray,roundcorner=10pt]
\textbf{System role prompt:} \newline\newline
Your task is to correct the predicted SQL based on the provided feedback by expert human.\newline
\newline
1. Input Information: You will receive a question, a database schema, a predicted SQL query, and a human feedback.\newline
\newline
2. SQL format in your response:\newline
    - You must ensure that your response contains a valid SQL.\newline
    - The format of SQL in your response must be in the following format: ```sql SQL ```.\newline
\newline
\newline
\textbf{User role prompt:}\newline
\newline
- Schema Overview:  \{schema\} \newline
- Columns Description:  {columns Description}
\newline
- Question: \{question\}  \newline
\newline
- Predicted SQL: ```sql \{Incorrect SQL\} ```
\newline
- Expert Human Feedback: \{feedback\}
    \end{mdframed}
    \centering
    \caption{The predefined prompt for interaction of magic agent with correction agent.}
    \label{fig:correction_prompt_appendix}
\end{figure}
\begin{figure}[]
    \begin{mdframed}[backgroundcolor=verylightgray,roundcorner=10pt]
\textbf{System role prompt} \newline
You are a helpful AI assistant that manages other assistants.\newline
\newline
\newline
\textbf{User role prompt} \newline
Manager, please review the following prompt for the following agent and generate a revised prompt. \newline
So far you have revised the prompt for \{iteration\_number\} times.\newline
Agent description: This agent generates feedback based on the comparison between predicted and ground truth SQL queries.\newline
Previous output of agent that was not useful: \{agent\_outputs[-1]\}\newline
Previous prompt of agent that you should revise: \{agent\_prompts[-1]\}\newline
\newline
Revise prompt (Return the entire prompt so it can be directly pass to the agent):
    \end{mdframed}
    \centering
    \caption{The prompt template of manager for revising the predefined prompts of feedback agent.}
    \label{fig:manager_revise_prompt_for_feedback_agent_appendix}
\end{figure}
\begin{figure}[]
    \begin{mdframed}[backgroundcolor=verylightgray,roundcorner=10pt]
\textbf{System role prompt:} \newline
You are a helpful AI assistant that manages other assistants.\newline
\newline
\newline
\textbf{User role prompt:} \newline
Manager, please review the following prompt for the following agent and generate a revised prompt. \newline
So far you have revised the prompt for \{iteration\_number\} times.\newline
Agent description: This agent generates corrections for SQL queries based on expert human feedback.\newline
Previous output of agent that was not useful: \{agent\_outputs[-1]\}\newline
Previous prompt of agent that you should revise: \{agent\_prompts[-1]\}\newline
\newline
\#\#\#\# 
\newline
The ideal output of agent should be the following but we cannot directly give the ideal output to the agent:\newline
Ideal output of agent: \{Correct SQL\}\newline
\#\#\#\#
\newline
\newline
Revise prompt (Return the entire prompt so it can be directly pass to the agent):
\newline
    \end{mdframed}
    \centering
    \caption{The prompt template of manager for revising the predefined prompts of correction agent.}
    \label{fig:manager_revise_prompt_for_correction_agent_appendix}
\end{figure}
\begin{figure}[ht]
    \begin{mdframed}[backgroundcolor=verylightgray,roundcorner=10pt]
\# Guideline format:\newline
number. Reminder of mistake\newline
   - Question: "Question"\newline
   - Incorrect SQL generated by me: ```sql incorrect sql ```\newline
   - Corrected SQL generated by me: ```sql corrected sql ```\newline
   - Negative and strict step-by-step ask-to-myself questions to prevent same mistake again: \newline
\newline
\# Guideline so far:\newline
\{current\_guideline\} // current guideline variable is empty at first\newline
\newline
\# Recent mistakes that must be aggregate to Guideline:\newline
\{batch\_of\_successful\_feedbacks\}\newline
\newline
\newline
\# Updated Guideline (Return the entire of guideline):
    \end{mdframed}
    \centering
    \caption{The prompt of manager for self-correction guideline generation.}
    \label{fig:guideline_generation_prompt_appendix}
\end{figure}

\end{document}